\documentclass[letterpaper, 10 pt, conference]{ieeeconf}  

\IEEEoverridecommandlockouts                              

\usepackage{graphicx} 
\usepackage{mathptmx} 
\usepackage{amsmath} 
\usepackage{amssymb}  
\usepackage{newtxmath}
\usepackage{algorithm}
\usepackage{algpseudocode}
\usepackage{booktabs}
\usepackage{url}
\usepackage[hidelinks]{hyperref}
\usepackage{cite}
\newcommand{\qq}{\mathbf{q}}

\title{\LARGE \bf
Differentiable  Reinforcement Learning for Path Tracking by an Agile Fish-Like Robot
}

\author{Prashanth Chivkula$^{1}$, Kartik Loya$^{1 }$, Venkata Ravindhra Reddy Varikuti$^{1}$ and Phanindra Tallapragada$^{1}$
\thanks{$^{\dagger}$ All authors have contributed equally.
* This work was supported by grant 2433150 from the National Science Foundation. $^{1}$ The authors are with the Department of Mechanical Engineering, Clemson University, USA. {\tt\small email: \{ptallap\}@clemson.edu}}
}

\begin{document}
\maketitle
\thispagestyle{empty}
\pagestyle{empty}

\begin{abstract}

Fish-like swimming has inspired the design of several dozens if not hundreds of bioinspired robots in the last few decades. But the control and motion planning of such robots has been challenging due to the poorly modeled fluid-structure interaction and the nonlinear underactuated dynamics of such robots. While reinforcement learning has allowed significant advances in the context of ground and aerial robots, the lack of a suitable simulation environment with appropriate computational speed and accuracy have prevented similar progress for fish-like robots. We address this two-fold problem by developing a simulation platform that approximates the motion of our fish-like robot with computational efficiency. Then the motion control and path tracking by the robot is performed using PID control where the (variable) gains are learned using back propagation through time and training on a curriculum. The policy learned in the simulation  is then applied on the physical platform, demonstrating an excellent match.

\end{abstract}

\section{INTRODUCTION}
The locomotion of fish has several characteristics like speed, efficiency, agility and stealth that engineers have sought to emulate and this motivation has inspired the design of a variety of fish-like robots, such as \cite{barrett_mit_1996, Triantafyllou_ScientificAmerican_1995, pettersen_ieee_ram_2016,  zhu_science_2019} to name a few. Fish achieve high speed, efficiency and agility by exploiting the interaction of their body with the vortex wake and the added mass effect, \cite{lighthill_jfm_1970, Childress1981}. This fluid-structure interaction can be difficult to model or simulate in a fast and computationally efficient manner. So while dozens or hundreds of fish-inspired robots have been designed, the control and motion planning of such robots has remained a challenge. 

Reinforcement learning and data driven methods offer a means to address this challenge and advances in their application to fish-like robots have begun to be made, see for example \cite{xie_ifac_2020, mitsuhiro_icra_2021, murphy_tor_2021, rt_srep_2023, Z_wang2025}. Deep reinforcement learning (DRL) combined with central pattern generators and a purely data driven surrogate model have been used to learn to track a path by a fish like robot in \cite{xie_ifac_2020}. DRL along with a simple model of three link swimmer simulated in Mujoco were used to control the motion of a soft fish like robot with dielectric elastomer actuator in  \cite{mitsuhiro_icra_2021}. In \cite{Z_wang2025} the hydrodynamic model  of a robotic fish with pectoral fins was learned using bidirectional long short-time memory networks and its gaits were controlled using central pattern generators. In \cite{murphy_tor_2021} a Koopman operator was constructed using data from a physical model that was then used to control its motion. The motion control in these papers is limited to tracking a reference heading angle and thereby a path. The speed of swimming is usually very small, less than a fraction of a body length per second, a regime where the complexity of hydrodynamic effects are reduced. Furthermore the robot does not simultaneously track a speed.

In this paper, we describe a deep reinforcement learning framework for controlling the speed and heading angle of a fish like robot. Since models and efficient simulation platforms are absent for generating training data, we first develop a model and simulation environment for a planar carangiform type of fish-like robot, based on the models developed in \cite{ rt_srep_2023, tallapragada_ACC2015, tallapragada_jcnd2016, pft_nd_2019}. The parameters in these models are estimated using experimental data. The physical platform is a highly efficient and agile swimming robot whose primary means of propulsion is an internal rotor \cite{pollard2017, wiens2018slender, Free2020BioinspiredPW,chivkula2025}. The robot has two actuators, one for propulsion and one primarily for steering; this motivates the use of separate PID controllers for tracking a reference angle and one for tracking a speed. Reference angle tracking is used in conjunction with
a line-of-sight planner to track a path. The gains for these PID controllers however have to change with the curvature along the path and the reference speed \cite{sharma_2025}.
Recent work has shown that differentiable physics simulators enable efficient gradient-based learning of control policies by allowing gradients to propagate through system dynamics \cite{song2024learning, xu2022accelerated, xu2022efficient}. 
Since we develop a differentiable simulator of the Chaplygin sleigh dynamics, gradients can be propagated directly through the system model during training. The control policy (for these gains) is therefore learned using backpropagation through time on a curriculum of tracking tasks. The policy learned in simulation is then applied on the physical platform and is demonstrated by the robot tracking various curved paths at changing reference speeds.
See the attached supplementary file for a video and source codes are available at this Github repository (\href{https://github.com/kloya03/agilefish-diff-rl}{agilefish-diff-rl}).

\section{ROBOT DESIGN} \label{sec:design}
We developed a fish-like robot with a simple and modular architecture inspired by previously reported internally actuated rotor-driven systems \cite{tallapragada_ACC2015,Free2020BioinspiredPW}. Multiple variations of such inertial-propulsion platforms have been proposed in the literature \cite{chivkula2024hopping,wiens2018slender}, demonstrating the viability of oscillatory internal mass actuation for aquatic locomotion. Inspired by these platforms, we developed a modular and simplified variant designed specifically as a control-oriented experimental testbed. 
\begin{figure}[!ht]
      \includegraphics[width=0.5\textwidth]{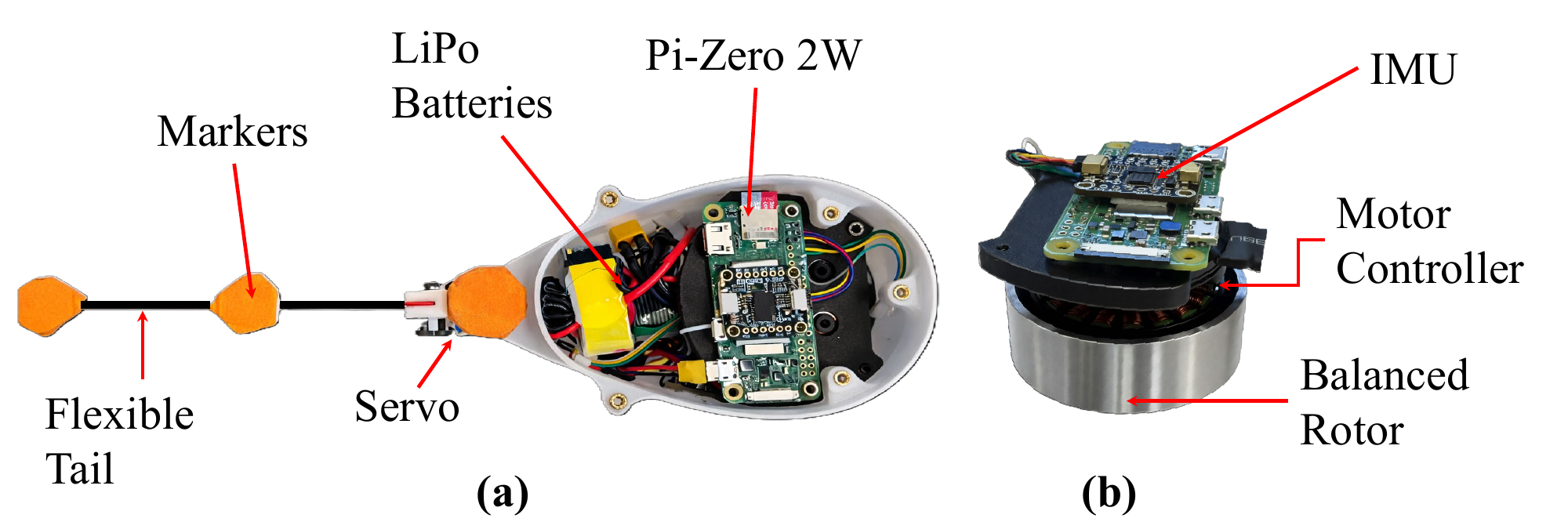}.
      \caption{(a) shows the top view of the robot. (b) shows the internal rotor module along with the electronics that consist of Pi Zero 2w and the IMU}
      \label{fig:robot_design}
\end{figure}

The robot consists of a rigid extruded shell shaped as a hydrofoil, referred to as the head, as shown in Fig.~\ref{fig:robot_design}(a). The head is $150$ mm long with an extruded height of $75$ mm, and the tail is $100$ mm long. The total length of the robot is $250$ mm (1 body length), and the total weight is $535$ g.
This sealed shell houses the propulsion module and onboard electronics, see Fig. \ref{fig:robot_design}(b). Propulsion is generated by an internally mounted balanced rotor that produces periodic inertial excitation, which interacts with the surrounding fluid to generate net forward thrust through internal momentum exchange rather than external propeller action. 
The rotor is driven by a brushless motor (370 Kv Vertiq 4006 module), enabling precise control of oscillation amplitude and frequency. Forward velocity is regulated by commanding a sinusoidal rotor velocity, $\dot{\phi} = A \omega sin(\omega t)$, where $A$ and $\omega = 2\pi \cdot f$. Onboard computation performs estimation and control, with an IMU (BNO085) providing inertial measurements for state estimation. A Raspberry Pi Zero 2W serves as the computer on board.
Heading control is achieved through a servo-actuated tail rudder. 
The robot achieves a minimum turning radius of 0.3 body lengths (BL) and a maximum swimming speed of 2 BL/s.
\section{MODELING AND SIMULATION } \label{sec:aigym}

\subsection{Mathematical Model}
\begin{figure}[h]
      \includegraphics[width=0.5\textwidth]{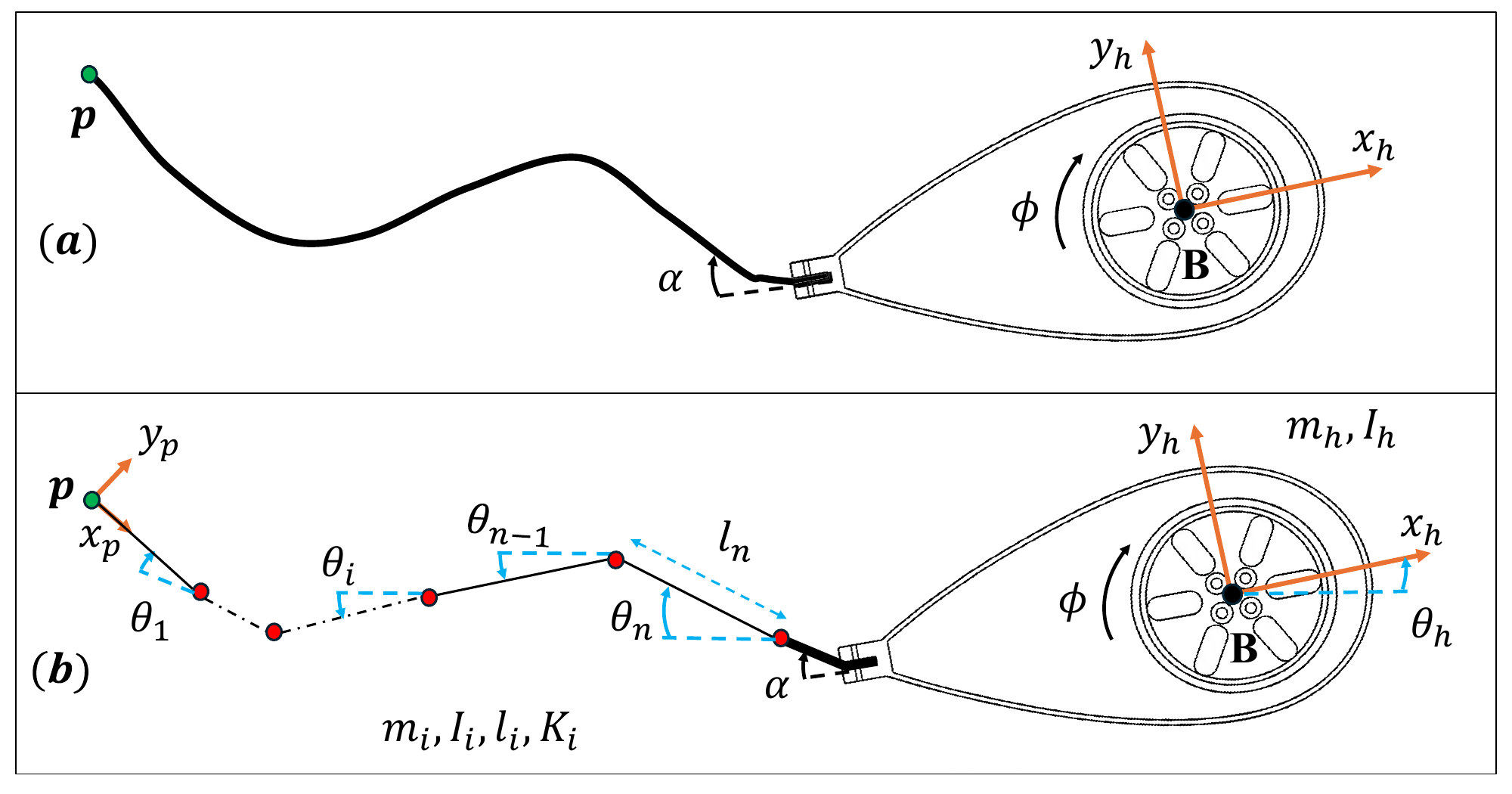}.
      \caption{(a) A Chaplygin sleigh with a heavy head at point B (center of mass) and a flexible tail link with a non-holonomic constraint at point C. (b) The flexible tail link is approximated as ’n’ multiple inextensible rods attached using torsional springs at the red points.}
      \label{fig:2d_multilink_sleigh}
\end{figure}

We construct a mathematical model of planar swimming for the robot described in section \ref{sec:design}. The purpose of the model is not detailed fluid reconstruction, but to capture the essential propulsion mechanism, steady-state limit-cycle behavior, and tail-induced rudder mechanism for steering. Propulsion in the robot is generated by an internal rotor mounted within the head, see Fig. \ref{fig:robot_design}b and  \ref{fig:2d_multilink_sleigh}. 

The robot's configuration is described by the planar pose of the head $( x_h, y_h, \theta_h ) \in SE(2)$ and the absolute orientation of each tail segment $\theta_i \in S^1$. Hence, the generalized coordinates for the system are $\qq = \begin{bmatrix}
   x_h, y_h, \theta_h, \theta_n, \cdots , \theta_2, \theta_1 
\end{bmatrix}$. Here, the rudder angle $\alpha(t)$ is actuated by a lightweight servo and is treated as a prescribed input rather than a generalized coordinate. 

 The kinetic energy of the robot is $T = \frac{1}{2}\dot{\qq}^T M(\qq,\alpha)\dot{q} + I_r \dot{\phi}^2$ where, $I_r$ is the rotor moment of inertia and  the mass matrix $M(\qq)$ includes both the mass and  the hydrodynamic added mass of each tail segment, \cite{lighthill_jfm_1970}.
Tail flexibility is modeled using torsional springs between consecutive links which gives the potential energy as $
    V = \frac{1}{2} k_j \Big( \sum_{j=1}^{n-1} (\theta_j-\theta_{j+1})^2 \Big) = \frac{1}{2} \mathbb{K}\mathbf{q}^{\intercal} \Tilde{A} \mathbf{q} $
Then, the Lagrangian $\mathcal{L} = T-V$ is given by,
\begin{equation}
    \mathcal{L} =  \frac{1}{2} \dot{\qq}^{\intercal} \mathcal{M}(\qq,\alpha) \dot{\qq} - \frac{1}{2} \mathbb{K} \qq^{\intercal} \Tilde{A} \qq
\end{equation}

Hydrodynamic viscous drag on the head and tail links is modeled using a Rayleigh dissipation function defined in the body frame of each rigid element. 
The viscous drag is modeled using the body-fixed frame velocities of each bodies $v_{b,i}$ and diagonal damping matrices $C_h$ for the head and $C_l$ for all the links. Then the Rayleigh dissipation function is defined as,
\begin{equation}
    \mathcal{R} = \frac{1}{2}v_{b,h}^T C_h v_{b,h} + \frac{1}{2} \sum_{i =1}^n  v_{b,i}^T C_h v_{b,i}
\end{equation}
Propulsion in oscillatory swimmers is dominated by the net momentum exchange generated by the oscillating rotor–tail system. Explicitly resolving the associated vortex shedding and wake evolution would require a fully coupled fluid–structure simulation, which is computationally expensive and complex for control-oriented modeling. Instead, we adopt a reduced-order mechanical representation that captures the dominant propulsion mechanism through added-mass effects and a nonholonomic constraint that enforces zero lateral velocity at an effective contact point $p$ located at the tail end. At this point, lateral slip is assumed to be suppressed, leading to the Chaplygin-type Pfaffian constraint
\begin{equation}\label{eq:non-holonomic}
    \mathcal{A}(\qq,\alpha)\dot{\qq} = 0
\end{equation}
In the inviscid flow approximation, such nonholonomic constraints have been shown to model the Kutta-Joukowski vortex shedding condition \cite{tallapragada_ACC2015,tallapragada_jcnd2016} and have inspired underwater robot design and control, see \cite{pollard2017, Free2020BioinspiredPW}. With the Lagrangian $\mathcal{L}$, the generalized damping forces $Q_d = -\frac{\partial \mathcal{R}}{\partial \dot{\qq}}$, and the nonholonomic constraint at point $p$ defined in \eqref{eq:non-holonomic} the equations of motion are obtained by combining the Euler–Lagrange equations with the constraint using a Lagrange multiplier. This yields

\begin{align}\label{eq:Euler_Lag}
  \frac{\mathrm{d}}{\mathrm{d}t}\Big(\frac{\partial \mathcal{L}}{\partial \dot{q}_i }\Big)-\frac{\partial \mathcal{L}}{\partial q_i} + \frac{\partial{\mathcal{R}}}{\partial{\dot{\qq}}} = \mathcal{A}^T(\qq,\alpha)\dot{\qq} \lambda,\\
  s.t.~~~ \mathcal{A}(\qq,\alpha)\dot{\qq} = 0
\end{align}

The resulting constrained equations of motion are written as,
\begin{equation} \label{eq:model}
    \begin{bmatrix}
        M(\qq,\alpha) & B(\dot{\qq},\qq,\alpha) \\
        \mathcal{A}(\qq,\alpha) & 0
    \end{bmatrix} \begin{bmatrix}
        \ddot{\qq} \\ \dot{\qq}
    \end{bmatrix}  = \begin{bmatrix}
        \mathbb{K} \Tilde{A} \qq + Q_d(\dot{\qq},\qq,\alpha) + \mathcal{A}(\qq,\alpha) \lambda \\ 0
    \end{bmatrix}
\end{equation}
This formulation captures the primary hydrodynamic forces while maintaining a compact structure suitable for analysis, fast simulation and learning-based control.

\subsection{Validation} \label{sec:valid}
To assess the fidelity of the reduced-order model, we compare its predictions against experimental data collected from the physical robot. The damping and torsional stiffness parameters were identified from $12$ experimental trials, each of $20$ seconds duration.

\begin{figure}[h]
    \includegraphics[width=0.48\textwidth]{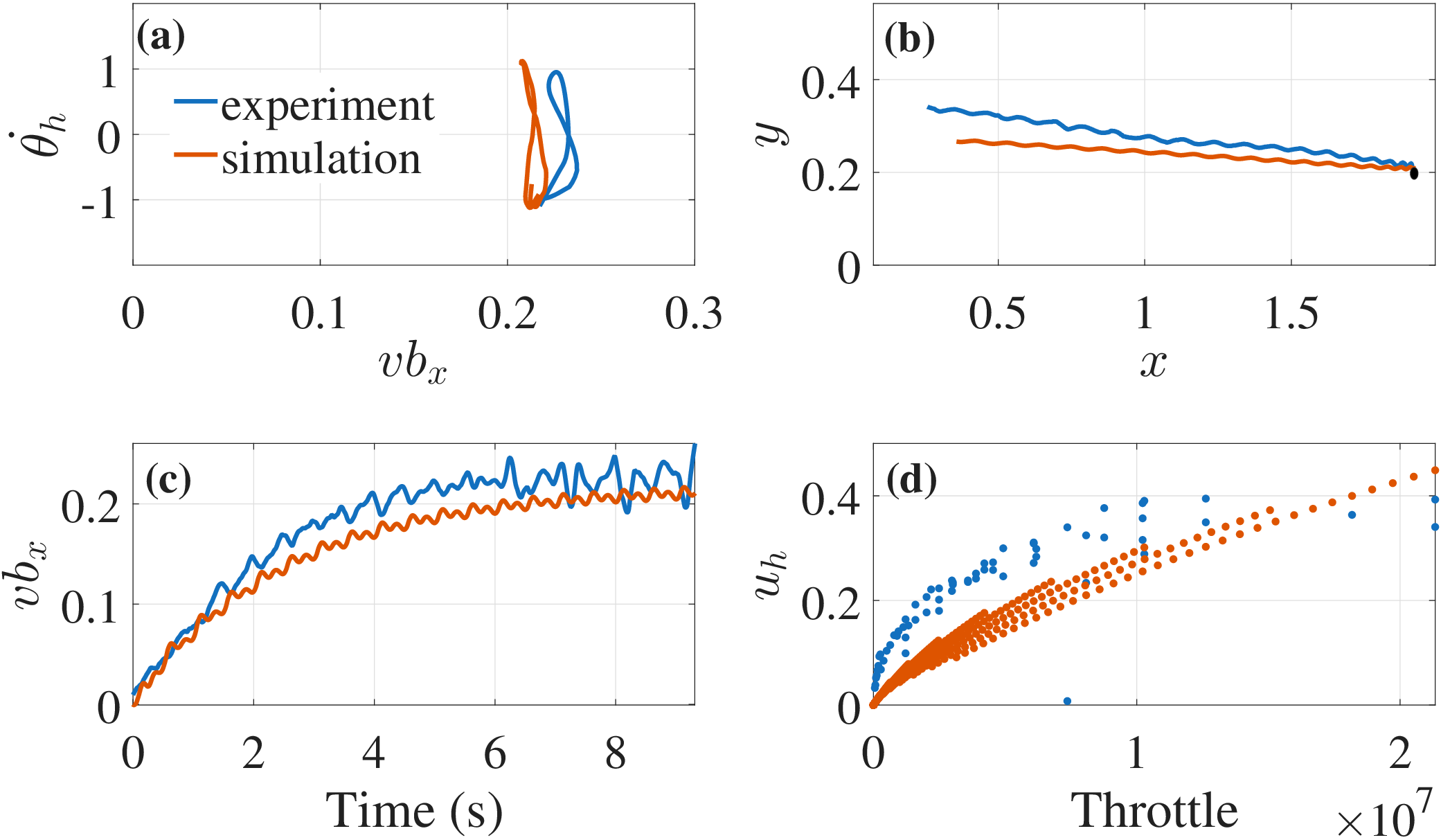}.
      \caption{Experimental and simulated validation of the reduced-order model for the representative case A=9, $f$=2. (a) Phase portraits show matching limit-cycle behavior. (b)$X-Y$ path trajectory. (c) Body fixed frame forward velocity $(v_{bx})$ comparison. (d) steady-state forward velocity scales approximately linearly with throttle $(s)$ input $(A\cdot \omega^2)^2 $ }
      \label{fig:validation}
\end{figure}
Figure \ref{fig:validation} compares simulation and experimental results for a representative actuation case (amplitude $A = 9$ rad and frequency $\omega = 2 \pi \cdot 2$). The forward velocity (let $v_{bh,x} := v_{bx}$ for ease of notation) response shows strong agreement in both transient behavior and steady-state magnitude. The nonlinear model reproduces the stable limit-cycle dynamics induced by oscillatory actuation, and the resulting steady propulsion accurately matches the experimentally observed planar $X-Y$ trajectory.
Experimental validation in Figure. \ref{fig:validation}(d) shows that with the presence of a stable limit-cycle, the steady-state forward velocity scales approximately linearly with a scalar throttle variable. Over the operating range of $0.1–0.3$ m/s, velocity exhibits an approximately linear relationship with both the amplitude of acceleration ($A\omega^2$) and the square of it, that is $(A\omega^2)^2$, with the latter providing a slightly better empirical fit. We therefore define the throttle input as,
\begin{equation}
    s := A^2\omega^4
    \label{eq:throttle_map}
\end{equation}
This limit-cycle structure suggests regulating motion by shaping oscillatory behavior rather than directly commanding forces. Thus, the robot’s long-term speed is governed by the limit-cycle behavior and can be regulated by adjusting the actuation parameters, also shown in \cite{pft_nd_2019}.

\subsection{Control Formulation}
The control objective consists of simultaneous forward velocity tracking and geometric path tracking through heading regulation. Propulsion depends on oscillation amplitude ($A$) and frequency ($f$), while steering is achieved through rudder deflection ($\alpha$), leading to coupled actuation channels. Manual PID tuning in this setting is nontrivial because speed and heading dynamics interact through the underlying limit-cycle behavior, requiring significant heuristic adjustment to achieve stable and responsive performance. We therefore adopt a structured dual-loop PID controller: one loop regulates heading via $\alpha$ using heading error ($e_{\theta}$), 
\begin{equation}\label{eq:PID}
    \hat{\alpha} = K_{p,a} e_{\theta} + K_{i,a} \int e_{\theta} \mathrm{d}t + K_{d,a} \frac{\mathrm{d}e_{\theta}}{\mathrm{d}t}
\end{equation}
and the other regulates forward speed via the throttle variable $s$, which is informed by limit cycles. 
\begin{equation}\label{eq:PID}
    s = K_{p,s} e_v + K_{i,s} \int e_v \mathrm{d}t + K_{d,s} \frac{\mathrm{d}e_v}{\mathrm{d}t}
\end{equation}
Here, $e_v$ is the error in velocity. Instead of selecting gains heuristically, the PID gain vector is treated as a set of decision variables to be optimized.


\begin{figure*}[ht]
    \centering
    \includegraphics[width=0.85\textwidth]{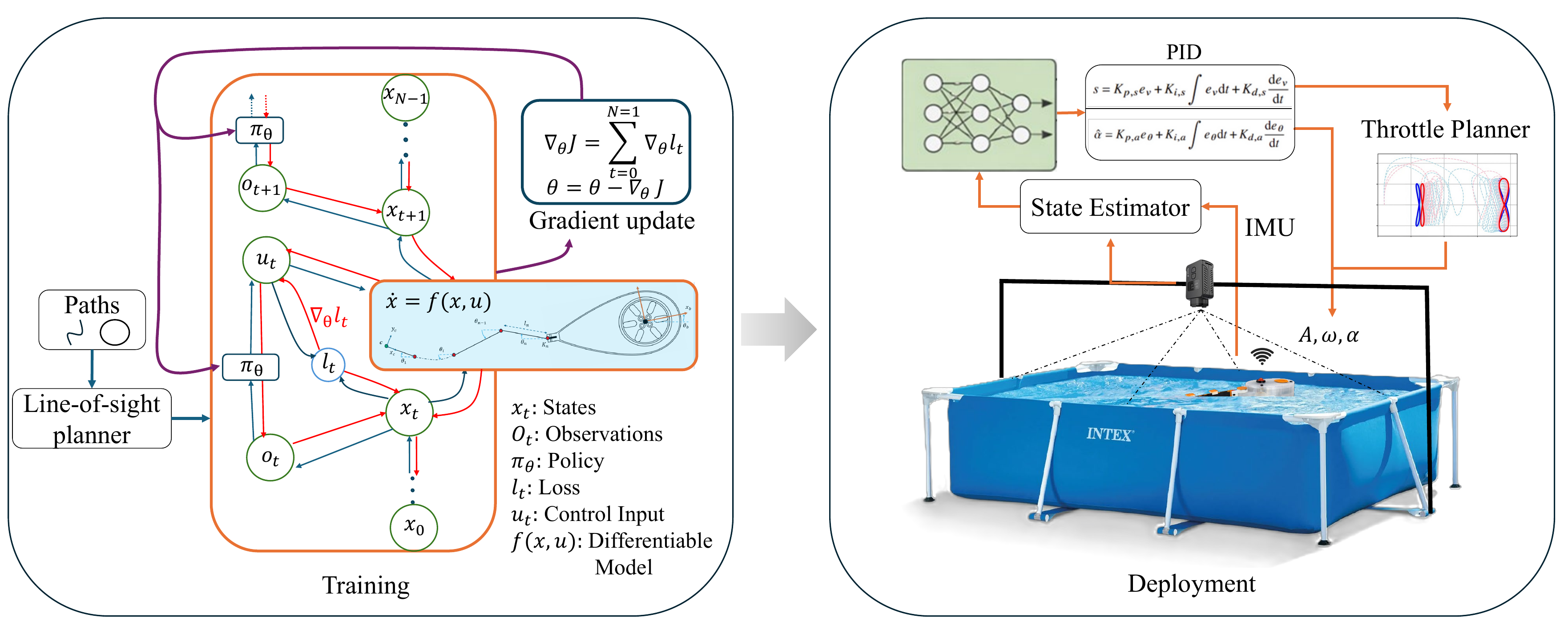}
    \caption{Learning and deployment pipeline. (Left) During training, PID gains are learned using differentiable reinforcement learning via backpropagation through time (BPTT) on the modified Chaplygin sleigh simulator with a line-of-sight path planner. (Right) The learned policy is deployed on the physical fish robot, where the network outputs adaptive PID gains for heading and speed control using state estimates from onboard sensing and camera feed.
    }
    \label{fig:path_tracking}
\end{figure*}

\section{Gain Optimization via Differentiable Closed-Loop Control}

The continuous time solutions to \eqref{eq:model} are time-discretised and represented as
\begin{equation}
x_{k+1} = f(x_k, u_k),
\end{equation}
where $x_k \in \mathcal{X}$ denotes the system state and $u_k \in \mathcal{U}$ the control input at time $t_k$. Observations are given by $o_k = g(x_k)$.
Instead of directly outputting control actions, a neural network parameterized by $\theta$ produces time-varying controller gains $K_k = \pi_\theta(o_k)$,
where $K_k$ collects the steering and throttle gains at time step $k$.
These gains parameterize the structured controller, which generates the control input $u_k = \mathcal{P}(x_k, K_k)$, where $\mathcal{P}$ denotes the PID control law  given in Eq. \eqref{eq:PID} with gains $K_k$ .

The resulting closed-loop system is
\begin{equation}
x_{k+1} = f\big(x_k, \mathcal{P}(x_k, \pi_\theta(g(x_k)))\big),
\end{equation}
making the entire trajectory implicitly dependent on $\theta$.

\subsection{Trajectory Optimization Objective}

We formulate controller learning as the finite-horizon optimal control problem.
\begin{equation}
\min_{\theta} J(\theta),
\end{equation}
where,$J(\theta) = \sum_{k=0}^{N-1} \ell(x_k, u_k).$

The stage cost is defined as
\begin{equation}
\begin{split}
\ell_k =\;
& w_1 e_{\theta,k}^2
+ w_2 e_{v,k}^2
+ w_3 e_{cte,k}^2 \\
& +\; w_4 \Delta u_k^2
+ \lambda_{K} \|K_k\|^2
+ \lambda_{\Delta K} \|\Delta K_k\|^2 .
\end{split}
\end{equation}
\noindent
where, $e_{\theta,k}$ denotes the heading (steering) error, $e_{v,k}$  the body-frame forward velocity tracking error, and $e_{cte,k}$ is the cross-track error relative to the reference path. The term $\Delta v_k = v_k - v_{k-1}$ denotes the change in body-frame forward velocity, for smooth changes in velocity. The gain vector at time step $k$ is given by $K_k$, where $\|K_k\|^2$ penalizes large gain magnitudes and $\|\Delta K_k\|^2 = \|K_k - K_{k-1}\|^2$ penalizes rapid temporal variations in the gains.

\subsection{Gradient Computation via BPTT}
The PID gains are generated by a neural policy
\(
K_k = \pi_\theta(o_k), 
\)
which parameterizes the closed-loop controller.
We minimize the finite-horizon objective
\begin{equation}
J(\theta) = \sum_{k=0}^{N-1} \ell(x_k,u_k),
\end{equation}
subject to the closed-loop dynamics
\(
x_{k+1} = f(x_k,u_k),
\quad
u_k = \mathcal{P}(x_k,K_k).
\)
Applying the chain rule over the unrolled trajectory yields
\begin{equation}
\nabla_\theta J
=
\sum_{k=0}^{N-1}
\left(
\frac{\partial \ell_k}{\partial x_k}
\frac{\partial x_k}{\partial \theta}
+
\frac{\partial \ell_k}{\partial u_k}
\frac{\partial u_k}{\partial \theta}
\right),
\end{equation}
where state variations are recursively induced by the system dynamics. The gradient flow through the closed loop can be summarized as
\begin{equation}
\theta 
\;\rightarrow\;
K_k 
\;\rightarrow\;
u_k 
\;\rightarrow\;
x_{k+1} 
\;\rightarrow\;
\ell_k .
\end{equation}

Gradients are computed by unrolling the closed-loop system for $N$ steps and applying backpropagation through time (BPTT)~\cite{song2024learning}, allowing differentiation through the PID update, throttle mapping, and physical dynamics.
The parameters are updated using gradient descent:
\begin{equation}
\theta \leftarrow \theta - \alpha \nabla_\theta J.
\end{equation}

\section{REINFORCEMENT LEARNING FOR GAIN SELECTION}
\subsection{Policy Architecture}
We next summarize the policy architecture and optimization settings used to train the controller through the differentiable closed-loop simulation.
At each time step $k$, the policy receives an input vector $z_k \in \mathbb{R}^{6}$ composed of normalized tracking signals and including the normalized head angle and velocity.
\begin{equation}
z_k =
\big[
\theta_{h,k}^{\mathrm{norm}},\;
v_{bx,k}^{\mathrm{norm}},\;
v_{\mathrm{des}}^{\mathrm{norm}},\;
e_{\theta,k}^{\mathrm{norm}},\;
e_{v,k}^{\mathrm{norm}},\;
e_{cte,k}^{\mathrm{norm}}
\big]^\top ,
\end{equation}
\noindent
where $\theta_{h,k}^{\mathrm{norm}}=\theta_{h,k}/(\pi/2)$, $v_{bx,k}^{\mathrm{norm}}=v_{bx,k}/v_{\max}$, $v_{\mathrm{des}}^{\mathrm{norm}}=v_{\mathrm{des}}/v_{\max}$, $e_{\theta,k}^{\mathrm{norm}}=e_{\theta,k}/(\pi/2)$, $e_{v,k}^{\mathrm{norm}}=e_{v,k}/v_{\max}$, and $e_{cte,k}^{\mathrm{norm}}=e_{cte,k}/L_d$.
The policy outputs a vector $a_k \in \mathbb{R}^{6}$, which is mapped to bounded PID gains using an elementwise $\tanh$ squashing followed by affine scaling:
\begin{equation}
K_k = \mathrm{scale}(\tanh(a_k)),
\end{equation}
where $K_k = [K_{p,a}, K_{i,a}, K_{d,a}, K_{p,s}, K_{i,s}, K_{d,s}]^\top$.

In particular, the steering gains are constrained to $K_{p,a},K_{i,a},K_{d,a}\in[0,1]$, and the speed gains are constrained to $K_{p,s}\in[0,2.5]$, $K_{i,s}\in[0,0.1]$, and $K_{d,s}\in[0,0.1]$.
Given the gain vector $K_k$, two PID updates are applied in closed loop. The steering PID uses the heading error $e_{\theta,k}$ to produce a steering command $\alpha_k$, followed by a rate limiter to enforce actuator slew limits. In parallel, the speed PID uses the velocity error $e_{v,k}$ to produce a throttle signal $s_k \in [0,1]$.
The throttle signal $s_k$ is transformed into motor amplitude and frequency parameters $(A_k,\omega_k)$ via a deterministic mapping as in  \eqref{eq:throttle_map}. These parameters define the rotor phase and angular acceleration according to
\begin{align}
\varphi_{k+1} &= \mathrm{mod}\!\left(\varphi_k + \omega_k \Delta t,\; 2\pi\right), \\
\ddot{\phi}_k &= -A_k\,\omega_k^{2}\,\sin(\varphi_{k+1}),
\end{align}
yielding the control inputs $(\alpha_k,\ddot{\phi}_k)$ applied to the sleigh dynamics.

Policy parameters are optimized using Adam with global-norm gradient clipping (1.0), using BPTT with an unroll length of $N=15$ time steps per update, similar to \cite{song2024learning}.
Each policy update is computed over a batch of $B=8$ independent environments. The policy is parameterized by a fully connected multilayer perceptron (MLP) with three hidden layers of sizes $128$, $128$, and $64$, each followed by ReLU activations.
Training was conducted on a multi-core CPU on Clemson's Palmetto Cluster \cite{antao2024modernizing}.

\subsection{Curriculum for Training}
Training is performed entirely in simulation using procedurally generated reference trajectories. We employ a curriculum that varies both commanded speed and path curvature to progressively increase task difficulty, similar to prior work \cite{li2026scaling, narvekar2020curriculum, chivkula2022curriculum}. Within each batch, all environments track a common base reference trajectory, while the commanded speeds are set deterministically to span $v_{bx}\in[0.1,0.3]$. 
This exposes the policy to diverse speed-tracking conditions while preserving a consistent path tracking objective across the batch.

In addition to speed variation, we adopt a curvature-based progression similar to \cite{chivkula2022curriculum}. Training begins with low-curvature trajectories and gradually increases path difficulty over time. The policy alternates between sinusoidal and circular reference paths every second episode. For sinusoidal trajectories, the amplitude (1.0 m and 1.5 m) and number of cycles (0.5, 1, and 2) are varied to increase curvature and lateral acceleration demands. For sinusoidal trajectories, a small random phase offset (uniformly sampled over $[0, 2\pi]$)  is also applied independently to each environment.
This preserves curvature magnitude while preventing the policy from memorizing a fixed spatial pattern.
For circular trajectories, the radius is reduced between 1.5 m and 1.0 m to impose tighter turning requirements.
Progression to higher difficulty levels is performance-driven: the curriculum advances only when the mean cross-track error over a sliding window of recent episodes falls below predefined thresholds (0.03 m and 0.01 m). This ensures that increased curvature is introduced only after the policy demonstrates adequate tracking performance at the current level. Together, these variations in speed and curvature encourage the policy to first develop stable speed and heading regulation on smoother, slower trajectories before adapting to tighter turns, higher lateral accelerations, and more demanding cross-track tracking conditions.

\section{Simulation Results}

\begin{figure}[t]
    \centering
    \includegraphics[width=1\linewidth]{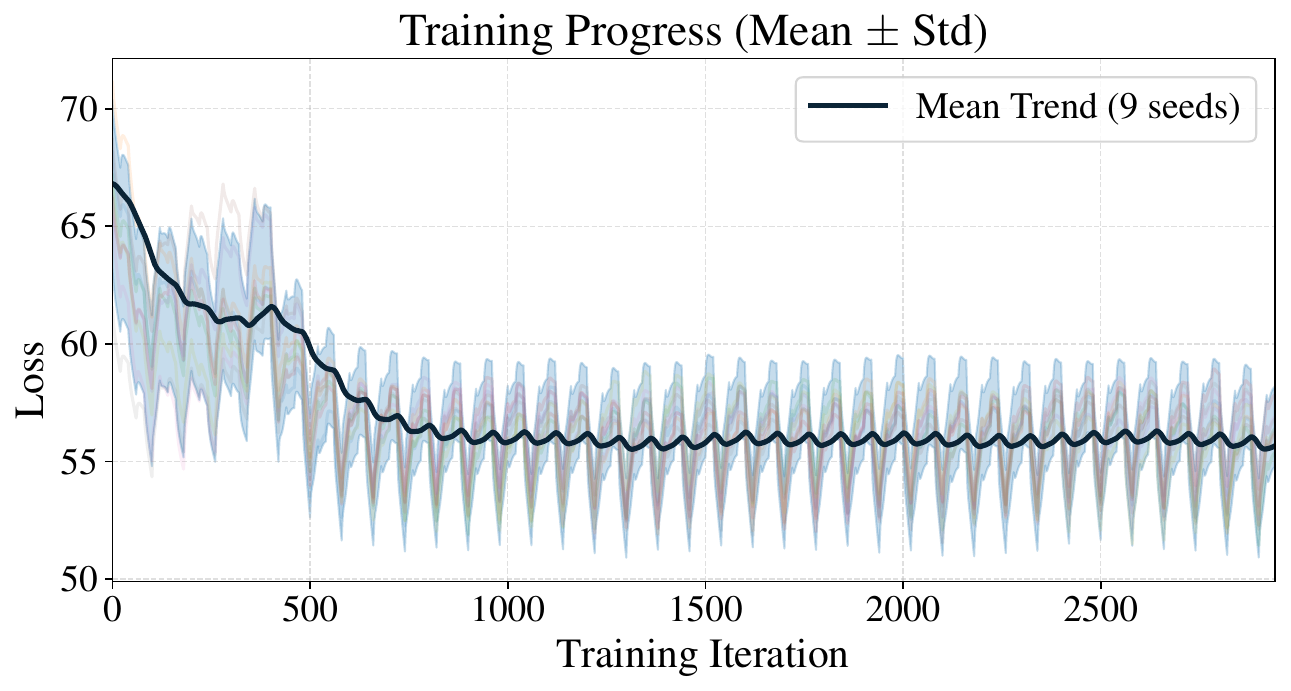}
    \caption{BPTT training curve (mean $\pm$ std) over 9 seeds. The bold line shows the mean loss, and the shaded region indicates one standard deviation.}
    \label{fig:tbptt_training}
\end{figure}

\begin{figure}[h!]
  \centering
  \includegraphics[width=\linewidth]{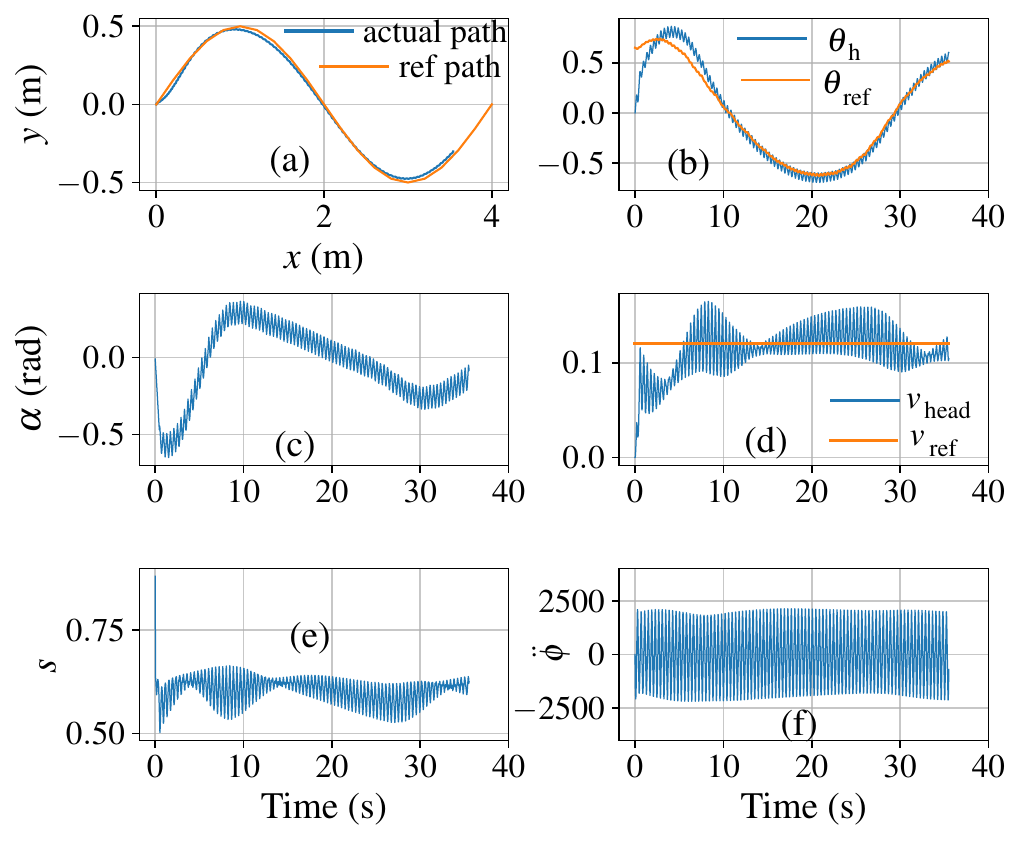}
  \caption{(a) shows the actual path (in blue) traced by the robot’s center of mass while tracking a sinusoidal reference trajectory of amplitude $0.5\,\mathrm{m}$. (d) shows the body velocity $v_{b,x}$ tracking the reference speed (orange) of $0.12\,\mathrm{m/s}$. (c) and (e) present the steering and throttle commands, $\alpha$ and $s$, respectively, as a result of the gains given by the trained policy described in Fig \ref{fig:gains_sine}, while (f) shows the resulting rotor angular acceleration $\ddot{\phi}$ produced by these control actions.}
  \label{fig:simulation_sine}
\end{figure}

\begin{figure}[t]
  \centering
  \includegraphics[width=\linewidth]{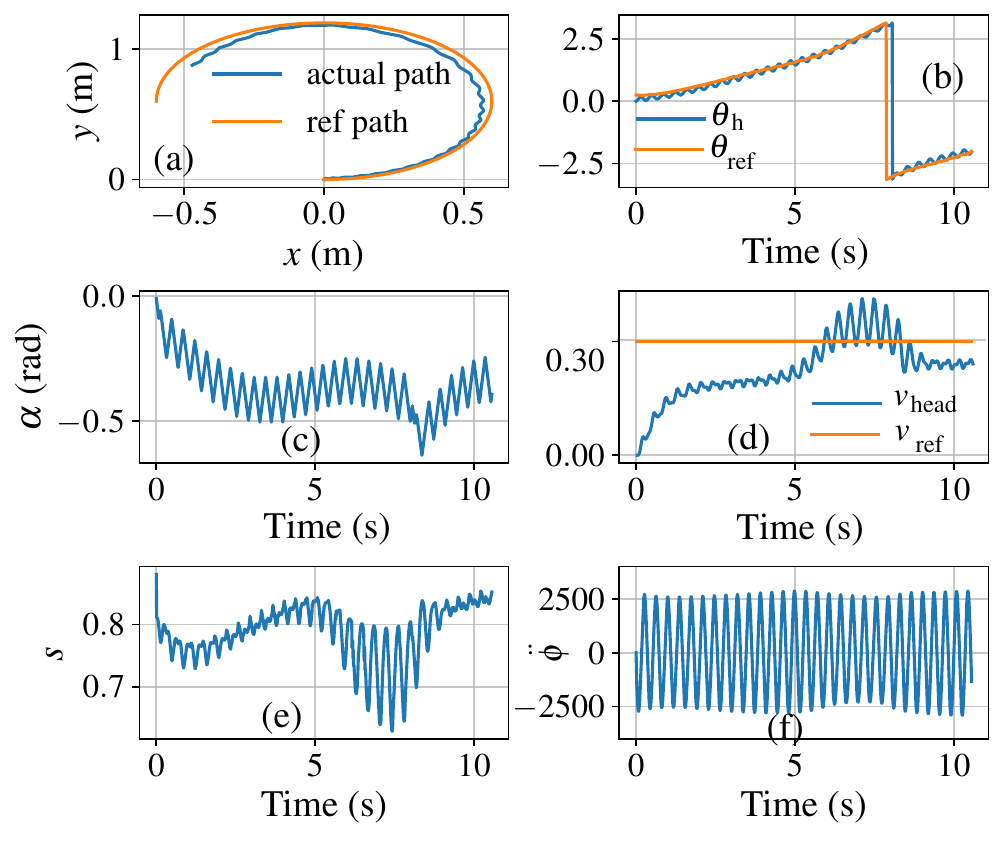}
  \caption{(a) shows the actual path (in blue) traced by the robot’s center of mass while tracking a circular reference trajectory of radius $0.6\,\mathrm{m}$. (d) shows the body velocity $v_{b,x}$ tracking the reference speed (orange) of $0.3\,\mathrm{m/s}$. (d). (c) and (e) present the steering and throttle commands, $\alpha$ and $s$, respectively, as a result of the gains given by the trained policy described in Fig \ref{fig:gains_circle}, while (f) shows the resulting rotor angular acceleration $\ddot{\phi}$ produced by these control actions.}
  \label{fig:simulation_circle}
\end{figure}

To assess the effectiveness of the proposed learning framework, we evaluate the closed-loop tracking performance of our trained policy in simulation across varying curvature levels and commanded speeds $v_{bx} \in [0.1,\,0.3]$. 
Fig. \ref{fig:tbptt_training} shows the BPTT training curve averaged over nine random seeds. The mean loss decreases rapidly during the initial training phase, indicating that the policy quickly learns regulation of heading and speed. The periodic variations in the loss correspond to the alternating curriculum between sinusoidal and circular trajectories. Despite these transitions, the overall loss trend decreases and stabilizes. The narrow standard deviation band across seeds suggests consistent convergence behavior.
As shown in Figs.~\ref{fig:simulation_sine} and \ref{fig:simulation_circle}, the controller achieves a mean cross-track error of approximately $0.01\,\mathrm{m}$ across both sinusoidal and circular trajectories while maintaining negligible heading error. Comparable tracking accuracy is obtained across additional evaluated trajectories, including straight-line and square-wave paths, which are omitted for brevity.

A small but persistent velocity tracking error of approximately $0.05\,\mathrm{m/s}$ is observed in the circular trajectory (Fig.~\ref{fig:simulation_circle}) when tracking the higher reference speed of $0.3\,\mathrm{m/s}$, whereas for the sinusoidal trajectory, a lower reference speed of $0.12\,\mathrm{m/s}$  was tracked with a mean velocity error below 0.01 m/s. Although the commanded velocity is constant along each reference path, maintaining a strictly constant speed is not necessarily dynamically feasible for a swimming robot subject to curvature-induced hydrodynamic constraints. In particular, tighter turns require greater turning effort, which can reduce the achievable forward velocity. The residual speed error, therefore, reflects an inherent trade-off between strict velocity tracking and accurate path following.

\begin{figure}[h!]
  \centering
  \includegraphics[width=\linewidth]{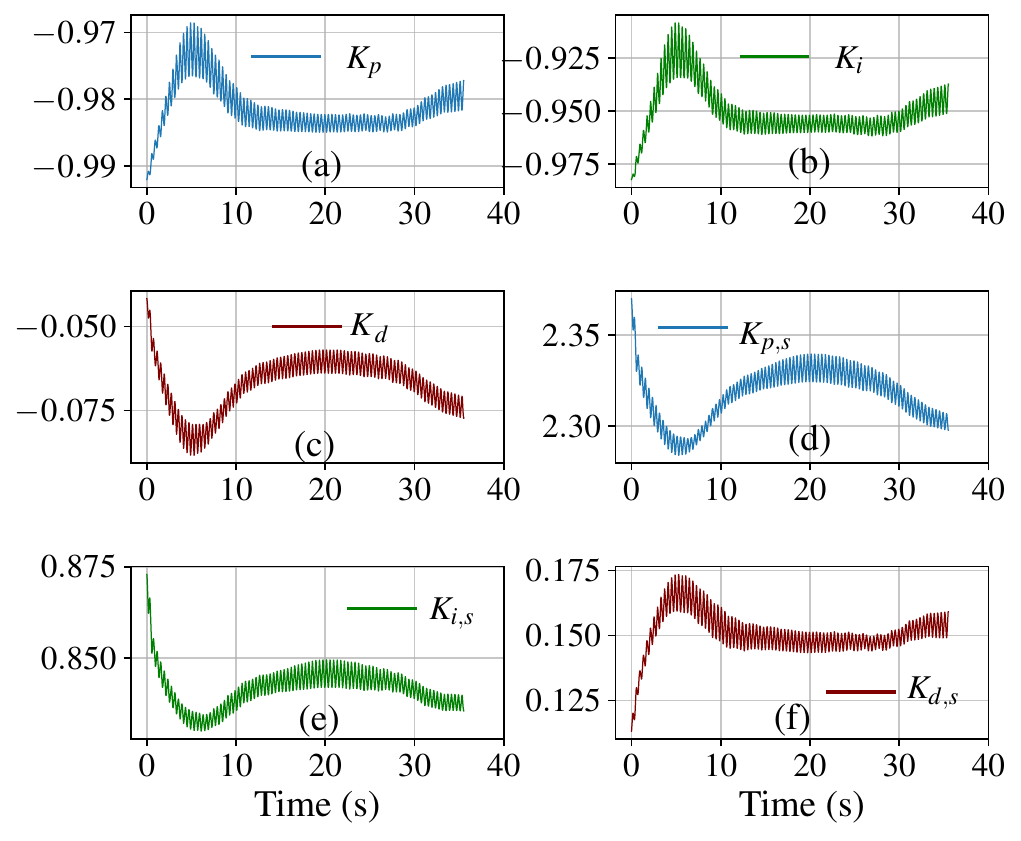}
  \caption{Subfigures (a)--(f) show the gains produced by the trained policy for the steering input $\alpha$ (see Fig \ref{fig:simulation_sine}(c)) and throttle $s$ (see Fig \ref{fig:simulation_sine}(e)) while tracking the sinusoidal reference path of amplitude 0.5 m in Fig \ref{fig:simulation_sine}(d).}
  
  \label{fig:gains_sine}
\end{figure}

Figure \ref{fig:gains_sine} illustrates the gains generated by the trained policy during tracking of a sinusoidal trajectory. The steering and speed gains vary smoothly over time, exhibiting approximately sinusoidal profiles that reflect curvature-dependent gain modulation. As the swimmer enters higher-curvature segments, the proportional and integral gains increase, while the derivative gain exhibits a slight decrease.
The reduction in derivative gain near curvature peaks indicates that the policy moderates damping when larger heading rates are naturally required by the path geometry, thereby preventing excessive resistance to the curvature-induced turning motion.

\begin{figure}[h!]
  \centering
  \includegraphics[width=\linewidth]{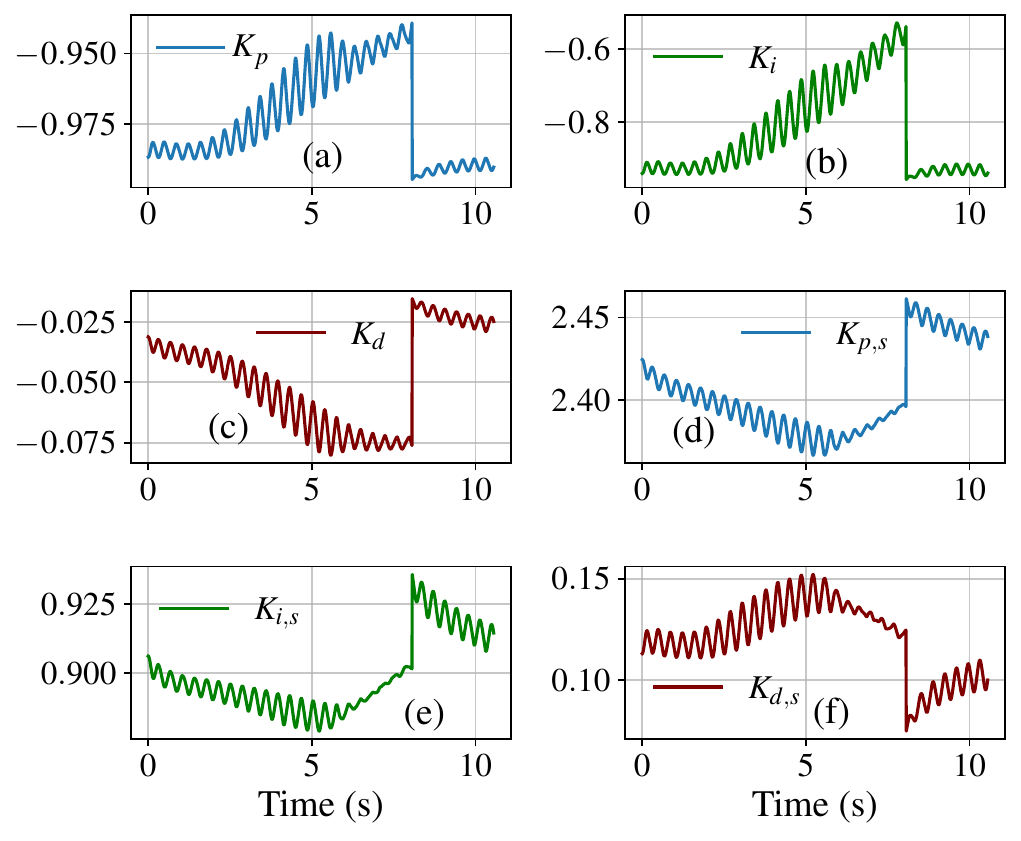}
  \caption{Subfigures (a)--(f) show the gains produced by the trained policy for the steering input $\alpha$ (see Fig \ref{fig:simulation_circle}(c)) and throttle $s$ (see Fig \ref{fig:simulation_circle}(e)) while tracking the circular reference path of radius $0.6\,\mathrm{m}$ in Fig \ref{fig:simulation_circle}(d) and a speed of $0.3$ m/s}
  \label{fig:gains_circle}
\end{figure}

The speed proportional gain exhibits a slight decrease near curvature peaks. Because turning motion and forward propulsion are dynamically coupled in undulatory swimming systems, temporarily relaxing speed regulation during high-curvature segments reduces interference with geometric path tracking. Although the gains vary systematically with curvature, their deviations from the nominal mean values remain small. These modest adjustments suggest that relatively minor geometry-dependent variations in the gains are sufficient to achieve accurate tracking along the sinusoidal path.

Similarly, during circular trajectory tracking in Fig \ref{fig:gains_circle}, the gains converge toward nearly steady values consistent with the constant curvature of the path. A small jump is observed around $7.5$ seconds, which corresponds to angle wrapping in the heading error as the swimmer progresses along the circle. This discontinuity is due to the heading transitioning across the $\pm \pi$ boundary rather than a change in curvature. 
Following this transition, the gains quickly resettle to values appropriate for the constant turning condition. 

Overall, the simulation results demonstrate that the learned policy achieves accurate path tracking while performing structured, curvature-dependent gain modulation. The consistent behavior across sinusoidal and circular trajectories indicates that the policy adapts directly to path geometry rather than exhibiting arbitrary time-varying tuning. 


\section{EXPERIMENTAL RESULTS}
A series of experiments were conducted to evaluate the real-world performance of the trained policy across different reference trajectories. The objective is to assess closed-loop tracking accuracy on the physical robotic fish  and examine whether the behavior observed in simulation translates to practical operation. The policy is deployed directly onboard on the robot, and performance is quantified using root-mean-square (RMS) cross-track, heading, and  speed errors.

\subsection{Experimental setup}

Experiments were conducted in a $14\,\text{ft} \times 7\,\text{ft}$ indoor pool equipped with an overhead vision system for global position feedback. Colored markers mounted on the robot body were detected using HSV-based segmentation and contour extraction. Pixel coordinates were corrected for lens distortion and converted to metric world-frame positions. To reduce measurement noise, low-pass filtering was applied, and vision measurements were fused with onboard IMU data using a Kalman filter to obtain consistent estimates of position, velocity, and orientation for closed-loop control.

Control was executed onboard a Raspberry Pi Zero 2W within a ROS-based architecture. Path tracking was implemented using a line-of-sight planner that selects a lookahead point on the reference trajectory at a fixed radius of $0.4\,\text{m}$ from the robot. A PID controller generated the control inputs for heading and forward speed, with gains produced online by the learned actor policy. This onboard deployment enabled evaluation of tracking performance and curvature-dependent gain adaptation under real hydrodynamic conditions.
\begin{figure}[ht]
    \centering
    \includegraphics[width=0.5\textwidth]{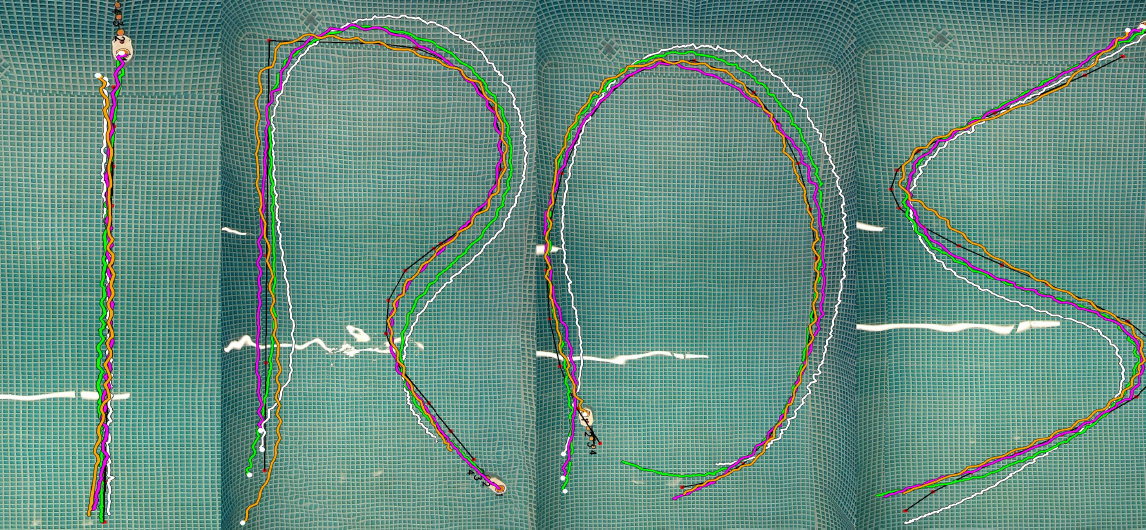}
    \caption{
       Path tracking performance of the fish robot at different forward 
    velocities (0.1, 0.15, 0.2, and 0.25~m/s). Trajectories are color-coded 
    by commanded speed, with white, green, magenta, and orange corresponding 
    to 0.1, 0.15, 0.2, and 0.25~m/s, respectively.
    }
    \label{fig:path_tracking}
\end{figure}

\begin{figure}[t]
    \centering
    \includegraphics[width=1\linewidth]{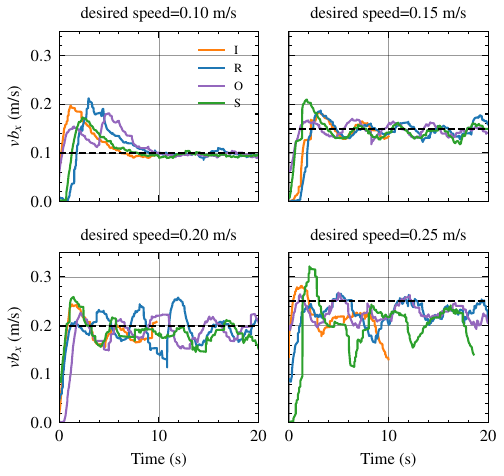}
    \caption{Forward velocity tracking performance across the I, R, O, and S reference 
paths at commanded speeds of 0.1, 0.15, 0.2, and 0.25~m/s . Solid curves denote smoothed trends of the measured body-frame forward velocity. The dashed 
horizontal lines indicate the commanded speeds.}
    \label{fig:velocity_tracking}
\end{figure}
\begin{figure}[!h]
    \centering
    \includegraphics[width=\linewidth]{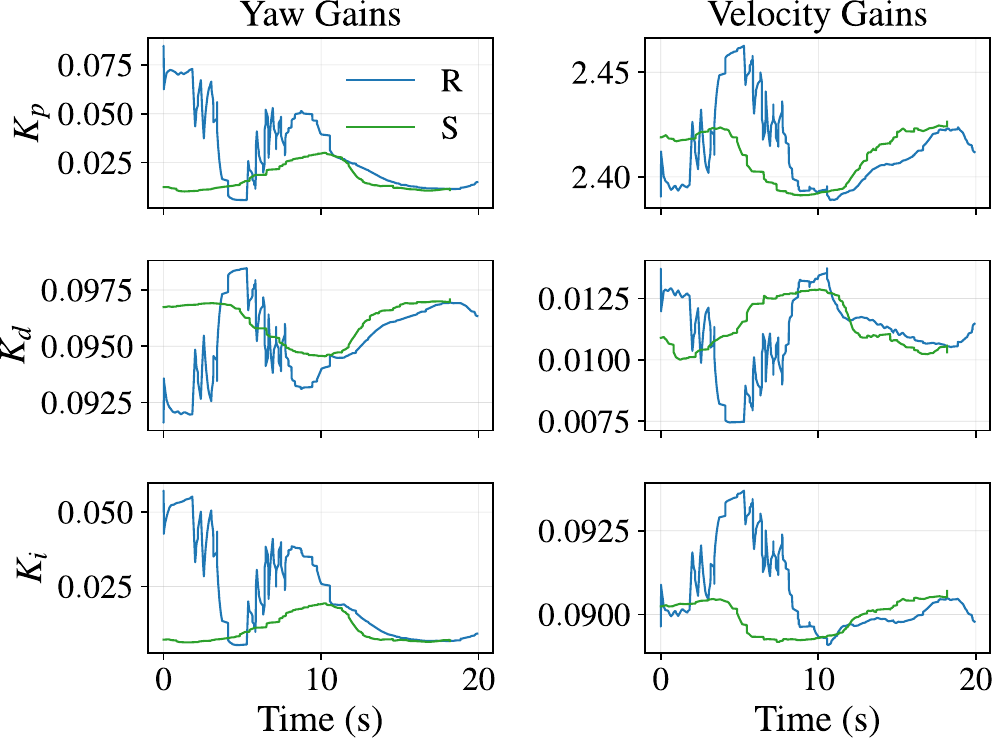}
    \caption{Online PID gains during experimental tracking of the ``R'' and ``S'' trajectories at 0.2\,m/s. 
Left: yaw gains. Right: velocity gains.}
    \label{fig:real_rs_gains}
\end{figure}
\subsection{Experimental tracking results}
The trained policy was evaluated on four reference trajectories forming the letters ``I'', ``R'', ``O'', and ``S'' at command speeds 
$v_{bx} \in \{0.10,\,0.15,\,0.20,\,0.25\}\,\text{m/s}$. 
These trajectories span varying curvature profiles, ranging from near-straight motion (``I'') to segments with tighter turns (``S'').

Across all trajectories and command speeds, stable closed-loop tracking was achieved with bounded cross-track and heading errors. Figure~\ref{fig:path_tracking} shows representative path overlays for different velocities. The swimmer successfully follows the geometric shape of each trajectory. Tracking accuracy exhibits a systematic dependence on command speed. Larger cross-track deviations are observed at lower velocities, while tracking improves as forward speed increases. This trend is consistent across the ``I'', ``R'', ``O'', and ``S'' trajectories. Quantitative RMS cross-track, heading, and speed errors for all trajectories and speeds are summarized in Table~\ref{tab:tracking_errors}.

The increased cross-track error at lower speeds can be attributed to the coupled training objective and the velocity-dependent steering authority of the fish robot. The policy is optimized to regulate both path tracking error and forward speed. Steering corrections reduce forward velocity, and at low command speeds aggressive heading adjustments can significantly disturb speed regulation. Consequently, the learned policy adopts a more conservative steering strategy to preserve velocity tracking, resulting in slightly increased lateral deviation.

Figure~\ref{fig:velocity_tracking} illustrates longitudinal velocity tracking for each trajectory. A small transient overshoot is observed at the beginning of each trial across all commanded speeds. Although a derivative term is present in the speed controller, the initial control action induces a short acceleration phase before feedback regulation settles the motion. The overshoot remains bounded and quickly converges to the commanded value, indicating stable closed-loop behavior. Figure~\ref{fig:real_rs_gains} shows the online evolution of PID gains during experimental tracking of the ``R'' and ``S'' trajectories at 0.2\,m/s. 
Yaw gains vary across trajectories and over time, whereas velocity gains remain comparatively stable with minimal modulation.

Overall, the experimental results demonstrate that the learned policy generalizes effectively to the physical robot. The controller maintains stable path tracking while performing consistent, curvature-dependent gain modulation across multiple trajectories and commanded speeds.

\begin{table}[!h]
\centering
\setlength{\tabcolsep}{5pt}
\renewcommand{\arraystretch}{1.15}
\begin{tabular}{c|cccc|cccc}
\toprule
& \multicolumn{4}{c|}{Cross-Track RMSE (m)} 
& \multicolumn{4}{c}{Speed RMSE (m/s)} \\
Path 
& \multicolumn{4}{c|}{Velocity (m/s)} 
& \multicolumn{4}{c}{Velocity (m/s)} \\
& 0.10 & 0.15 & 0.20 & 0.25 
& 0.10 & 0.15 & 0.20 & 0.25 \\
\midrule
I & 0.016 & 0.019 & 0.011 & 0.012 
  & 0.036 & 0.041 & 0.031 & 0.040 \\

R & 0.104 & 0.061 & 0.049 & 0.025 
  & 0.030 & 0.032 & 0.039 & 0.061 \\

O & 0.122 & 0.073 & 0.088 & 0.024 
  & 0.025 & 0.015 & 0.041 & 0.041 \\

S & 0.084 & 0.049 & 0.043 & 0.046 
  & 0.023 & 0.022 & 0.028 & 0.075 \\
\bottomrule
\end{tabular}
\caption{RMS tracking errors for I, R, O, and S paths at different forward velocities (m/s).}\label{tab:tracking_errors}
\end{table}

\section{CONCLUSIONS}
The paper presents results on the motion control and path tracking of a fish-inspired swimming robot using a reinforcement learning framework. 
A physics-based differentiable simulator with parameters identified from experimental data is used to enable gradient-based optimization of the control policy through backpropagation through time. The policy is structured as a gain-learning framework for PID controllers, thus maintaining an interpretable control structure.
The policy learned in simulation is transferred to the experimental platform where accurate simultaneous speed and path tracking is achieved at different reference speeds. This sim-to-real transfer is enabled by the mutlistep process of modeling and curriculum learning. While the results are limited to planar swimming, future work will address three-dimensional motion control.





\bibliographystyle{IEEEtran}
\bibliography{IEEEabrv,RL_PID}

\end{document}